\pdfoutput=1

\documentclass[11pt]{article}

\usepackage[final]{acl}
\usepackage{booktabs, multirow} 
\usepackage{soul}
\usepackage{xcolor,colortbl} 
\usepackage{changepage,threeparttable} 

\usepackage{times}
\usepackage{latexsym}
\usepackage{booktabs}
\usepackage{graphicx}  
\usepackage{amsmath} 
\usepackage{booktabs}      
\usepackage{arydshln}      
\usepackage{multirow}
\usepackage{placeins}
\usepackage[table]{xcolor} 
\usepackage{graphicx}      
\usepackage{amsmath}  
\usepackage{makecell}
\usepackage[abs]{overpic}
\usepackage{subcaption}
\usepackage{tcolorbox}
\usepackage[ukrainian,english]{babel}
\usepackage{multicol,lipsum}

\usepackage[T1]{fontenc}

\usepackage[utf8]{inputenc}

\usepackage{microtype}

\usepackage{inconsolata}
\usepackage{float}

\usepackage{stfloats} 
\usepackage{graphicx}

\title{Introducing OmniGEC: A Silver Multilingual Dataset for Grammatical Error Correction}

\author{Roman Kovalchuk \\
  Ukrainian Catholic University, Softserve \\
  Lviv, Ukraine \\
  \href{mailto:r.kovalchuk.pn@ucu.edu.ua}{r.kovalchuk.pn@ucu.edu.ua}
  \\\And
  Mariana Romanyshyn \\
  Grammarly \\
  Kyiv, Ukraine \\
  \href{mailto:mariana.romanyshyn@grammarly.com}{mariana.romanyshyn@grammarly.com}
  \AND
  Petro Ivaniuk \\
  Softserve \\
  Lviv, Ukraine \\
  \href{mailto:ivanyukpetro@gmail.com}{ivanyukpetro@gmail.com} \\}

\begin{document}
\maketitle 
\begin{abstract}
In this paper, we introduce OmniGEC, a collection of multilingual silver-standard datasets for the task of Grammatical Error Correction (GEC), covering eleven languages: Czech, English, Estonian, German, Greek, Icelandic, Italian, Latvian, Slovene, Swedish, and Ukrainian. These datasets facilitate the development of multilingual GEC solutions and help bridge the data gap in adapting English GEC solutions to multilingual GEC. The texts in the datasets originate from three sources: Wikipedia edits for the eleven target languages, subreddits from Reddit in the eleven target languages, and the Ukrainian-only UberText 2.0 social media corpus. While Wikipedia edits were derived from human-made corrections, the Reddit and UberText 2.0 data were automatically corrected with the GPT-4o-mini model. The quality of the corrections in the datasets was evaluated both automatically and manually. Finally, we fine-tune two open-source large language models — Aya-Expanse (8B) and Gemma-3 (12B) — on the multilingual OmniGEC corpora and achieve state-of-the-art (SOTA) results for paragraph-level multilingual GEC. The dataset collection and the best-performing models are available on Hugging Face\footnote{\url{https://huggingface.co/collections/lang-uk/omnigec-68095391ebef195ed6c0a5f3}}.

\end{abstract}

\section{Introduction}

\subsection{Motivation}

Grammatical Error Correction (GEC) is a task within Natural Language Processing (NLP) to identify and correct grammatical errors in written text. It is widely used in education, language learning, and professional communication. While researchers have made significant advancements in GEC for high-resource languages like English, its development for multilingual contexts remains an active research area. Most languages, including Ukrainian, Czech, Slovene, and others, remain underrepresented and understudied in GEC, lacking “golden” (high-quality, human-annotated) and “silver” (high-quantity, automatically annotated) datasets and methods that effectively account for the linguistic diversity and grammatical complexity of different languages.

The English GEC spearheaded advancements in GEC, and some of the developed methods and approaches can be directly applied to other languages. For instance, the authors of the recent survey paper \cite{pillars} mention that for ensembling and ranking the results, a high diversity between possible corrections results in higher scores. This approach can be applied and validated for a variety of languages. At the same time, many solutions are English-centric and unadjustable to other languages, creating language bias \cite{ban_english_nlp}. For example, the GECTOR model \cite{gector}, used for ranking the proposed grammatical corrections, is specifically trained to work with English, and its adaptation to other languages would be extremely high-effort.

With the introduction of transformer-based models \cite{transformers} and modern large language models (LLMs), the landscape in modern GEC shifted drastically \cite{gpt4_gec, chatgpt_or_grammarly}: (1) synthetic data generation has started to be used more often to rely less on high-quality parallel data \cite{text_simplification_by_tagging}, and (2) open-source LLMs opened new possibilities to approach the GEC task with various prompting and fine-tuning techniques \cite{pillars}. These models and methods have been successfully applied to the English language but have not been validated in the multilingual setting.


\subsection{Problem Setting}

The lag in multilingual GEC is due to several reasons. First, large, high-quality data in multiple languages is expensive and difficult to standardize, making it hard for models to generalize. Additional gaps include a lack of ablation studies on data quality versus quantity, cross-language transfers, minimal exploration of reinforcement-based methods, and persistently low state-of-the-art (SOTA) scores for low- and mid-resource languages \cite{gec_2025, ged_2023}. 

We aim to address these gaps by: (1) publishing a multilingual silver GEC dataset collection called OmniGEC, comprising human edits from Wikipedia\footnote{\url{https://www.wikipedia.org/}} and synthetically generated corrections of Reddit\footnote{\url{https://www.reddit.com/}} subreddits and UberText 2.0 social media corpus\footnote{\url{https://lang.org.ua/en/ubertext/}}, (2) conducting ablation studies on a per-dataset basis, revealing their impact on the model's performance across target languages, and (3) comparing model performance before and after Low-Rank Adaptation (LoRA) \cite{peft} fine-tuning on Aya-Expanse (8B) \cite{aya-expanse} and Gemma-3-12B-IT \cite{gemma3}.

The rest of the paper is organized into the following sections. Section 2 covers related work in the area of multilingual GEC. Section 3 describes the collection of the OmniGEC datasets and their characteristics. Section 4 dives into the quality evaluation of the OmniGEC datasets. Section 5 describes the experimental setup for training multilingual GEC models and the corresponding metrics. Section 6 provides the analysis of experimental results, including an ablation study. The paper ends with conclusions, limitations, and ethical considerations.

\section{Related Work}

\citet{Bryant_2023} provide a comprehensive historical overview of GEC approaches, from rule-based methods and machine learning classifiers for correcting a specific type of mistake to more recent techniques, such as using transformers and language models for generating a corrected output. This survey paper mentions the benefits of LLM-based data generation for low-resource GEC systems.

A more recent survey paper by \citet{pillars} covers contemporary approaches in the era of large language models and explores the performance of proprietary and open-source LLMs for the English GEC. They set new state-of-the-art performance for the English language by ensembling several LLM-based correction outputs.

A large body of research in the area of GEC comes from monolingual and multilingual GEC shared tasks. The most recent competitions include MultiGEC-2025 \cite{gec_2025}, the first shared task in multilingual grammatical error correction, MultiGED-2023 \cite{ged_2023}, the first shared task in multilingual grammatical error detection, and UNLP-2023 \cite{unlp_shared_gec_2023}, the first shared task in Ukrainian grammatical error correction.

The MultiGEC-2025 shared task featured twelve European languages and was organized into two tracks: (1) minimal, for systems producing minimally corrected texts, and (2) fluency, for systems that prioritize fluency and idiomaticity. The winning team in both tracks, minimal and fluency, was UAM-CSI \cite{staruch-2025-uam}. They used the Gemma-2 (9B) model \cite{gemma2} with two LoRA adapters per track, one-to-many languages. Interestingly, all participating teams used only one instruction template in English for all languages and obtained relatively low scores for low- and mid-resource languages. To compare, the winning UAM-CSI team scored 69.15 F\textsubscript{0.5}\textsuperscript{minimal} and 69.68 F\textsubscript{0.5}\textsuperscript{fluency} for the Ukrainian language, while the best solutions of the UNLP-2023 shared task showed 73.14 F\textsubscript{0.5}\textsuperscript{minimal} and 68.17 F\textsubscript{0.5}\textsuperscript{fluency} for the Ukrainian language on the same data.

The organizers of the MultiGEC-2025 shared task used a combination of various pre-existing manually annotated GEC corpora for the target languages. They published a comprehensive overview of the resulting MultiGEC dataset used in the shared task \cite{multigec_dataset}. The dataset is rather small, with 400 to 1,000 sample texts per language. The language-specific subcorpora vary in size, annotation, and sources of original texts, which makes the dataset inconsistent. The MultiGED-2023 competition used the same dataset but for fewer languages.

Although both high-quality and high-quantity datasets exist in English \cite{clang8, conll_2014, bea-2019}, multilingual GEC data is limited. Despite providing the best collection of manually annotated multilingual GEC data, the MultiGEC dataset is still insufficient for thorough LLM fine-tuning, preference optimization, and ablation studies for multilingual GEC.

\section{Data}

In this section, we describe the creation of the OmniGEC datasets that cover eleven languages: Czech, English, Estonian, German, Greek, Icelandic, Italian, Latvian, Slovene, Swedish, and Ukrainian. The language selection was based on the MultiGEC-2025 shared task for further data and model comparability.

For consistency in data-related measurements, we employ GPT-4o \& GPT-4o-mini's tokenizer \cite{gpt4o}, and for model-related technicalities, we use Gemma-3 and Aya-Expanse's tokenizers respectively.

\subsection{Corpus Composition}

OmniGEC contains three silver-standard GEC subcorpora:
\begin{itemize}
    \item WikiEdits-MultiGEC — Wikipedia edits for the eleven target languages;
    \item Reddit-MultiGEC — subreddits from Reddit in the eleven target languages with synthetically generated corrections;
    \item UberText-GEC — the Ukrainian-only UberText 2.0 social media corpus with synthetically generated corrections.
\end{itemize}

\textbf{WikiEdits-MultiGEC} is a small dataset of human error corrections made by Wikipedia contributors for our target eleven languages. These corrections were obtained using the official Wikipedia API and cover six months, from September 28, 2024, to April 17, 2025. We collected only the edits from the category \texttt{newcomer task copyedit} as this category usually contains small grammatical mistakes. These edits can be found at the \texttt{Special:RecentChanges} page on Wikipedia\footnote{\url{https://en.wikipedia.org/w/index.php?tagfilter=newcomer+task+copyedit&title=Special:RecentChanges}}, but only the last 30 days or 500 pages of changes are retained, whichever limit is reached first. Empirical observations indicated that running the code monthly to update the dataset does not result in any data loss for the target languages. 

Dataset creation included three main steps: (1)~collecting metadata for all recent Wikipedia pages that received edits across the target languages, (2) collecting all edits from each page, and (3) post-processing and filtering edits from Wikipedia-specific artifacts.

The average number of samples per language is 1.6K, resulting in 1.2M tokens in total. It is important to note that we artificially capped the number of samples for the English language to avoid promoting further bias towards the only high-resource language in the dataset.

The data collection code can be found on GitHub\footnote{\url{https://github.com/PetroIvaniuk/wikiedits-multigec}}. Additional information about the dataset is provided in Appendix \ref{sec:appendix-wikiedits-multigec}.


\textbf{Reddit-MultiGEC} is a large multilingual corpus of posts scraped from Reddit (13M tokens in total), automatically corrected using the approach described in Section \ref{sec:synthetic-correction}. We selected subreddits where the primary language of communication was one of our target languages. Additionally, for Icelandic, which is extremely low-resource, we included a subreddit dedicated to learning Icelandic, with posts in English and Icelandic. Data post-processing included two main steps: (1) we classified all samples with the langid\footnote{\url{https://github.com/saffsd/langid.py}} language classifier, keeping only samples written in our target languages, and (2) ran automated content moderation with the omni-moderation-2024-09-26\footnote{\url{https://platform.openai.com/docs/guides/moderation}} model to filter out potentially offensive posts. The highest fraction of censored posts was in Italian, with almost 20\% of posts flagged, and the lowest fraction of flagged posts was in Icelandic — 2.8\%. The resulting corpus contains texts on a variety of topics with diverse natural errors for our target eleven languages. This dataset can be extended in the future, as we capped the collection at 400 of the latest subreddits per language as of March 25, 2025. The data collection code for Reddit-MultiGEC can be found on GitHub\footnote{\url{https://github.com/r-kovalch/omnigec-data}}.

\textbf{UberText-GEC} is a 25\% subset of UberText 2.0 social media texts, scraped from Ukrainian Telegram (22M tokens, out of 110M total) \cite{ubertext}. It was automatically corrected using the approach described in Section \ref{sec:synthetic-correction}. This dataset will significantly contribute to future ablation study experiments and the GEC model for the Ukrainian language. 

The distribution of samples and token length per language for golden (MultiGEC-2025) and silver (OmniGEC) datasets can be found in Figure \ref{fig:corpus_data} and Figure \ref{fig:token-len} respectively
 (Appendix~\ref{sec:appendix-datasets-comparison}).

\subsection{Synthetic Grammatical Error Correction Generation}
\label{sec:synthetic-correction}

\begin{figure*}[t]
  \includegraphics[width=\textwidth]{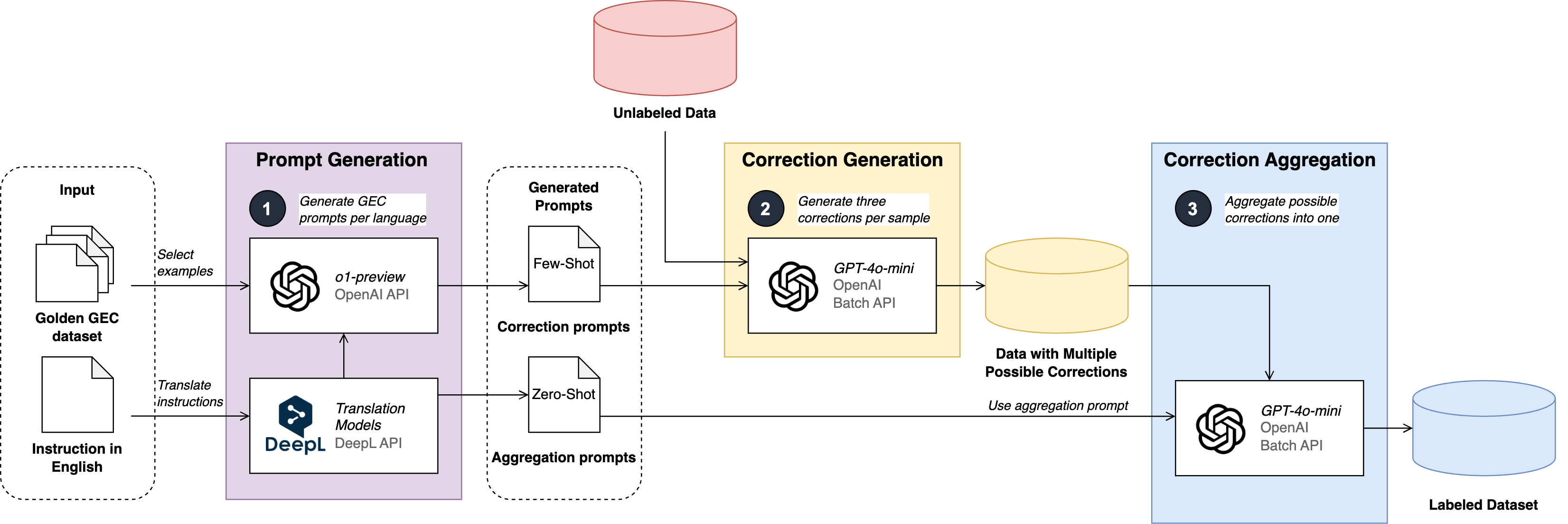}
  \caption{A schema for the three-step correction generation we followed for Reddit-MultiGEC and UberText-GEC.}
  \label{fig:synthetic-data-generation}
\end{figure*}

To generate grammatical error corrections, we employed DeepL\footnote{\url{https://www.deepl.com/}}, an AI-powered translation service that offers translations across 30 languages, and a two-stage LLM prompting approach with GPT-4o-mini and o1-preview \cite{o1}. The approach is visualized in Figure \ref{fig:synthetic-data-generation} and can be described in the following steps:

\begin{enumerate}
    \item Prompt Generation. First, we developed a GEC instruction in English and translated it into eleven target languages using DeepL. After that, for each language, we extracted correction examples from the development set of the MultiGEC dataset. We then prompted the o1-preview model to generate a few-shot prompt for each language based on the translated instruction and correction examples. The final few-shot prompts instruct the model to generate three possible grammatical error corrections.
    \item Correction Generation. For each language, we combined the few-shot prompts with paragraph-level raw text samples and prompted the GPT-4o-mini model to generate corrections for each sample.
    \item Correction Aggregation. Having obtained three corrections for each data sample, we prompted GPT-4o-mini again, instructing it to aggregate the corrections into one, creating a final correction. This aggregation prompt was also written in English and translated into eleven target languages with DeepL.
\end{enumerate}

The three-step correction generation approach is a slight variation of the high-diversity ranking and ensembling approach proposed in \cite{pillars}, as we aggregate multiple diverse corrections rather than selecting the best one. The reason behind this decision lies in the observation that even with low temperature, GPT-4o-mini "radiates" corrections into multiple possible outputs rather than having multiple complete corrections. Thus, aggregating them resulted in more complete corrections.

The prompting templates for all languages can be found on GitHub\footnote{\url{https://github.com/r-kovalch/omnigec-data}}.

%
%
%

\section{Quality Evaluation}

To assess the quality of corrections in the OmniGEC datasets, we used automated metrics and human feedback. We evaluated only the Ukrainian-language subcorpora due to time and human resource constraints and acknowledge the need for a further multilingual assessment. Nevertheless, we believe that the evaluation results still provide insights into the quality of corrections in the dataset.

For both evaluation tracks, we sampled 1,500 random examples from each of the three subcorpora, which totalled in 4,500 samples for evaluation.

\subsection{Automated Metrics}

Since we do not have golden human-annotated corrections to compare against, we generated reference corrections by three publicly available GEC systems: (1)~ Pravopysnyk \cite{bondarenko-etal-2023-comparative}, the UNLP-2023 shared task winner, (2) Spivavtor \cite{saini-etal-2024-spivavtor}, an instruction-tuned model for four text editing tasks in Ukrainian, including GEC, and (3) LanguageTool\footnote{\url{https://languagetool.org/}},  an open-source spelling and grammar checker for over 30 languages.

We then evaluated random 1,500 correction samples from each OmniGEC subcorpus (4,500 in total) against the three reference outputs with the ERRANT \cite{errant} and GLEU \cite{gleu_2015, gleu_2016, gleu_2016_referenceless} metrics, commonly used in GEC (see Table \ref{tab:multiref}). Such evaluation against multi-reference targets both provides insight into how aligned the corrections are with other systems' outputs and establishes a baseline for assessing future models.

\begin{table*}[t]
  \centering
  \footnotesize
    \begin{tabular}{lcccccc}
      \hline
      \textbf{Corpus} & \textbf{Precision} & \textbf{Recall} & \textbf{F\textsubscript{0.5}} & \textbf{GLEU} 
      & \textbf{Levenshtein Distance} 
      & \textbf{Character Error Rate}\\
      \midrule
            Reddit‑MultiGEC    & \textbf{17.92} & \textbf{59.51} & \textbf{20.84} & 46.89 & \textbf{36.87} & \textbf{18.20} \\
    UberText‑GEC       & 16.83          & 56.81          & 19.59          & 63.45          & 23.51          & 10.98          \\
      WikiEdits‑MultiGEC & 13.30          & 26.03          & 14.74          & \textbf{71.35} & 18.21          & 4.79           \\

      \hline
    \end{tabular}

  \caption{Multi-reference automated evaluation metrics across corpora with ERRANT (precision, recall, and F\textsubscript{0.5}), Levenshtein distance (error distance), character error rate (normalized error distance) and GLEU.}
  \label{tab:multiref}
\end{table*}

From Table \ref{tab:multiref}, we can see that with the increase in the character error rate (number of edits per 100 characters), the GLEU score decreases, and F\textsubscript{0.5} increases, which means that the more edits the corpus has, the lower GLEU score it yields in a multi-reference comparison.

\subsection{Human Evaluation}

The human evaluation of the OmniGEC corrections was set up as a grading task. We asked a pool of volunteers to grade the corrections on a scale from 1 to 5. The annotation instructions provided clear explanations and examples for each level of the scale. While complete annotation instructions are available on our GitHub\footnote{\url{https://github.com/r-kovalch/omnigec-data}}, we provide a brief explanation of the grades below:
\begin{enumerate}
    \item The correction introduced new errors, changed the meaning of the text, or changed the language.
    \item The corrected text contains major errors. 
    \item The corrected text is significantly improved over the original, but minor errors remain.
    \item The corrected text aligns with the Ukrainian orthography, a.k.a. the "minimal" grade.
    \item The corrected text aligns with the Ukrainian orthography and improves on fluency, a.k.a. the "fluency" grade.
\end{enumerate}

In total, 15 annotators participated in the project, all of whom were native speakers of Ukrainian. Most of the annotators were students majoring in linguistics. We received annotations for all 4,500 data samples, but only 100 samples were double-annotated due to time constraints.

\begin{figure*}[t]
  \includegraphics[width=\linewidth]{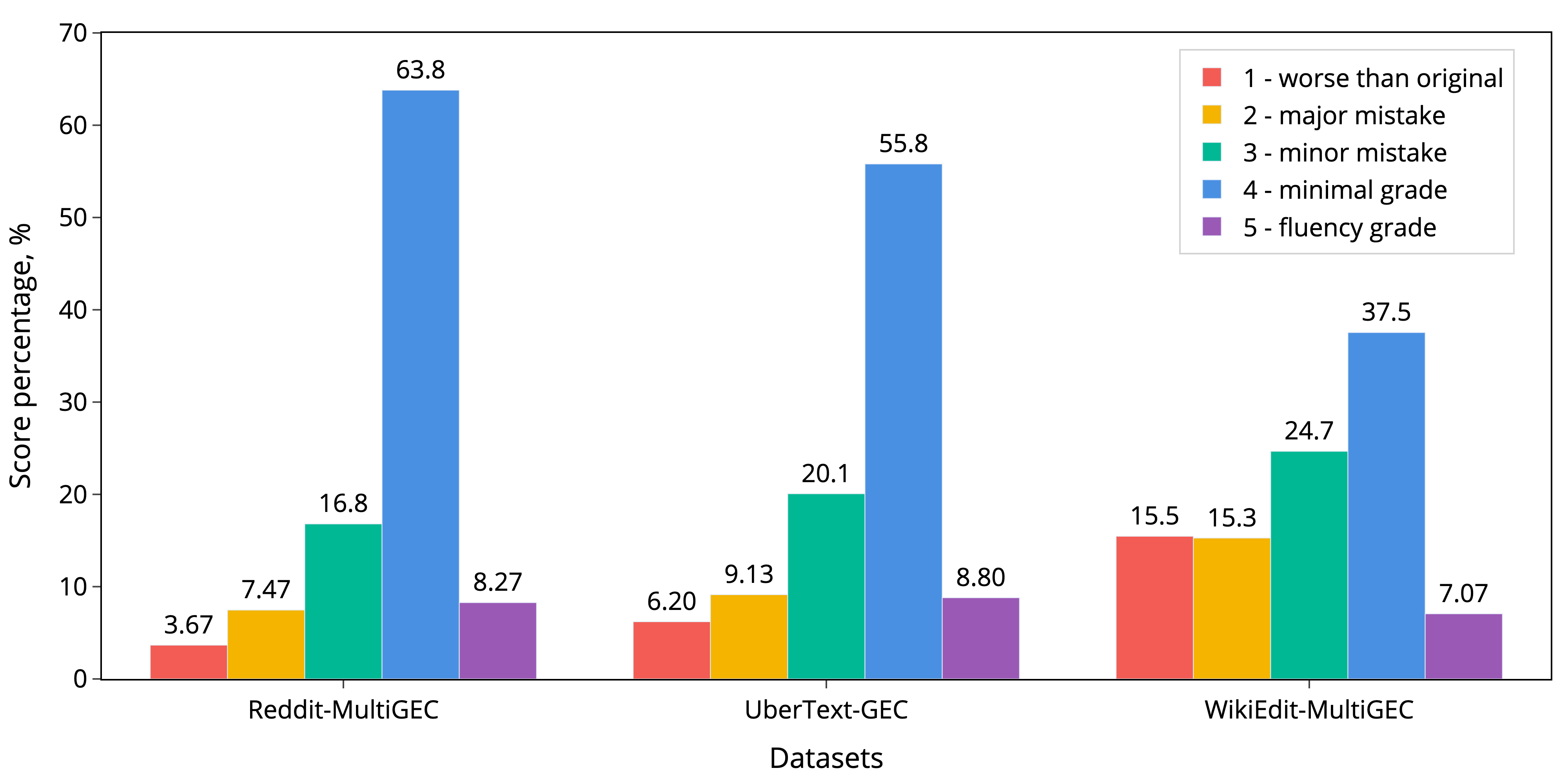}
  \caption{Grade distribution in human evaluations of corrections in OmniGEC datasets. The evaluation set contained 1,500 random Ukrainian-language samples from each subcorpus. }
  \label{fig:human-eval}
\end{figure*}

Figure \ref{fig:human-eval} shows the grade distribution across subcorpora. We observe that the extracted human-made corrections in WikiEdits-MultiGEC are of worse quality than the synthetically generated corrections in the other two subcorpora. The average grade in WikiEdits-MultiGEC is 3.05, while Reddit-MultiGEC and UberText-GEC average slightly higher, at 3.52 and 3.66, respectively.


The annotators also had an option to reject the sample if the original sentence was incomprehensible or the correction was impossible to judge. Only 2.8\% and 2.3\% of samples were rejected from Reddit-MultiGEC and UberText-GEC data, respectively, but the fraction of rejected samples in WikiEdits-MultiGEC was much higher, reaching~9.9\%.

\subsection{Error Analysis}

We conducted a manual error analysis to understand the primary causes of grades 1 and 2. Among the common issues present across all datasets were errors in the corrected texts, instances of overcorrection, and an excessive number of corrections within a single text, which made accurate evaluation challenging.

In addition to common errors, the low grades in \textbf{Reddit-MultiGEC} were used for non-ethical or inappropriate content, which was also rejected by annotators. In contrast, lower grades in \textbf{UberText-GEC} were largely due to additional non-essential text, such as promotional phrases like “subscribe to the channel” or “support us,” which negatively impacted the overall evaluation.

Grades 1 and 2 are the most prevalent in \textbf{WikiEdits-MultiGEC}, as shown in Figure \ref{fig:human-eval}. This led to a deeper investigation of the dataset to identify the root cause of the issue. The following causes were identified:

\begin{itemize}
\item\textbf{Information updates} — updates to dates, numbers, statistics, or records appear as text corrections in Wikipedia but are not grammatical error corrections.

\item\textbf{Domain-specific corrections} — annotators may lack domain knowledge to accurately grade edits in domain-specific texts.

\item\textbf{Distortion of context} — some samples contain excessive deletions of the input texts or large additions to the output texts.

\item\textbf{Data errors} — instances of poorly formatted text with embedded tags remain in the dataset, which can be fixed with more precise data cleaning.
\end{itemize}

For more details on the error types in the WikiEdits-MultiGEC refer to Appendix \ref{sec:appendix-error-analyis}.

\subsection{Overcorrection Bias in Generated Data}

Considering the nature of the three-step correction generation approach we employed for Reddit-MultiGEC and UberText-GEC, multiple correction aggregation makes the outputs subject to overcorrection. However, we consider that the benefits of the output being complete far overweight this bias, and the human evaluation study we conducted suggests that 70\%+ of examples are scored as "4 - minimal grade" and "5 - fluency grade", which we consider to be a good level of correction, especially for synthetically generated data. 

\section{Experiments}

In this section, we experiment with the OmniGEC dataset in the setting of the MultiGEC-2025 shared task.

\subsection{Model Choice}

Following the latest advancements, we focus on building an LLM-based GEC solution. We chose two open-source LLMs: Aya-Expanse (8B) and Gemma-3 (12B). Aya-Expanse has good target language coverage (5 out of 11), and its predecessor Aya-101 performed well in Ukrainian GEC \cite{saini-etal-2024-spivavtor}. Gemma-3 performs well in multilingual settings \cite{gemma3}, including in Ukrainian; however, the authors do not explicitly state which languages the model targets, other than "out-of-box" support for 35 languages and pre-trained support for over 140 languages. We chose the 12B version to examine the impact of parameter size in multilingual GEC, as both \citet{pillars} and \citet{aya_101} mention the sensitivity and non-linear improvements of size increase to the performance gained in GEC and multilingual tasks, respectively.

\subsection{Experimental Setup}

We conduct three incremental experiments for both the minimal and fluency MultiGEC-2025 tracks:
\begin{enumerate}
    \item MultiGEC — baseline, fine-tune the models solely on the MultiGEC train set.
    \item MultiGEC+Wiki — fine-tune the models on the MultiGEC train set and WikiEdits-MultiGEC.
    \item MultiGEC+Wiki+Reddit — fine-tune the models on the MultiGEC train set, WikiEdits-MultiGEC, and Reddit-MultiGEC.
\end{enumerate}

Due to time and cost limitations, we could not include UberText-GEC in our training experiments. We do include the fluency track — although our correction prompts targeted the minimal track, human annotations showed 7-9\% of examples with corrected fluency, so we evaluate the fine-tuned models against both tracks.

To evaluate our models and estimate the performance gained by adding the OmniGEC datasets, we use the GLEU score via the MultiGEC-2025 shared task CodaLab environment\footnote{\url{https://codalab.lisn.upsaclay.fr/competitions/20500}} and the MultiGEC-2025 test set.

The models are fine-tuned on paragraph-level data for better contextualization. We will, thus, be comparing our results with the best paragraph-level solution submitted to the MultiGEC-2025 shared task — Lattice \cite{seminck-etal-2025-lattice}, which was the second-best solution overall. The Lattice team fine-tuned LLaMA 3.0 (8B) \cite{touvron2023llamaopenefficientfoundation} for the task of paragraph-based multilingual GEC.

\section{Results}

In this section, we explore the results of our experiments, which include model performance in two MultiGEC-2025 tracks and the performance changes with the addition of OmniGEC training data.

\subsection{Baseline Overview}

\begin{table*}[ht]

\centering
\footnotesize
\begin{tabular}{lcccccc}
\hline
\textbf{Model} 
& \textbf{GLEU\rlap{\textsuperscript{mean}}\textsubscript{minimal}}
& \textbf{GLEU\rlap{\textsuperscript{mean}}\textsubscript{fluency}} 
& \textbf{GLEU\rlap{\textsuperscript{Ukrainian}}\textsubscript{minimal}}
& \textbf{GLEU\rlap{\textsuperscript{Ukrainian}}\textsubscript{fluency}}
& \textbf{GLEU\rlap{\textsuperscript{Estonian}}\textsubscript{minimal}} 
& \textbf{GLEU\rlap{\textsuperscript{Latvian}}\textsubscript{minimal}}
\\
\hline


\multicolumn{7}{l}{\textbf{Our Results}} \\
\noalign{\vskip 0.775mm}
\multicolumn{7}{l}{Aya-Expanse-8B} \\

\makecell[l]{
    \textit{MultiGEC}
}
    & 64.52
    & 48.37
    & \textbf{77.28}
    & 76.51
    & 33.27
    & 72.29
\\
\makecell[l]{
    \textit{MultiGEC+Wiki}
}
    & 65.16
    & 48.37
    & 77.05
    & \textbf{77.10}
    & 38.07
    & 73.04
\\
\makecell[l]{
\textit{MultiGEC+Wiki+Reddit}
}
    & 65.43
    & 49.80
    & 76.41         
    & 75.82
    & 41.52
    & 71.71
\\
\noalign{\vskip 0.775mm}
\multicolumn{7}{l}{Gemma-3-12B-IT}
\\
\makecell[l]{
    \textit{MultiGEC}
}
    & 61.43
    & 48.66
    & 74.25
    & 74.22
    & 54.74
    & 54.05
\\
\makecell[l]{
    \textit{MultiGEC+Wiki}
}
    & \textbf{67.02}
    & \textbf{52.34}
    & 75.17
    & 71.88
    & 55.12
    & \textbf{81.54}
\\
\makecell[l]{
    \textit{MultiGEC+Wiki+Reddit}
}
    & 66.42
    & 49.20
    & 75.11
    & 74.83
    & \textbf{57.54}
    & 80.19
\\
\midrule
\multicolumn{7}{l}{
    \textbf{MultiGEC-2025}
}
\\
\multicolumn{7}{l}{LLaMA-3-8B} \\
\makecell[l]{
    \textit{MultiGEC}
}
    & 56.85
    & -
    & 74.00
    & -
    & 44.02
    & 67.25
\\
\hline
\end{tabular}
\caption{The comparison of paragraph-based GEC models fine-tuned on the MultiGEC-2025 and OmniGEC datasets across all languages and specifically for Ukrainian, Estonian (minimal), and Latvian.}
\label{tab:ft-ukr}
\end{table*}

Table \ref{tab:ft-ukr} demonstrates the performance of fine-tuned models across all languages and specifically for Ukrainian, Estonian, and Latvian. For more detailed results per language, refer to Figure \ref{fig:face2face-minimal} and Figure \ref{fig:face2face-fluency} (Appendix~\ref{sec:appendix-training-results}).

Surprisingly, the 8B-parameter Aya-Expanse showed better baseline performance than the 12B-parameter Gemma-3. In the minimal track, it outperformed Gemma-3 for all languages except Estonian (Gemma-3 scored 21.47 more GLEU points than Aya-Expanse), Slovenian (2.42 more), and Swedish (7.46 more). However, it is worth noting that Aya-Expanse was not pre-trained to process these languages, and the ablation study in section~\ref{sec:ablation} shows that the quality generally increases with more data.

In the fluency track, Gemma-3 performed better on average despite being trained on fewer epochs than Aya-Expanse. For baseline training, we used early stopping on the validation dataset for both models. Only for Ukrainian, Aya-Expanse-8B scored almost two GLEU points more than Gemma-3 in fluency.

We presume that the small-sized Aya-Expanse benefited from a small golden MultiGEC dataset more than Gemma-3, as it requires fewer data for fine-tuning on downstream tasks and has much fewer excess languages: only 18 versus more than 100 supported languages in Gemma-3. At the same time, Gemma-3 has been trained on more languages, yielding a more uniform quality, even on the baseline, and outperforming the Aya-Expanse model on languages that Aya-Expanse does not support.




\subsection{Uniform Improvements}

Both Gemma-3 and Aya-Expanse yield better performance on average on both tracks when trained on both OmniGEC and MultiGEC data. Aya-Expanse's performance increased by 0.91 and 1.43 GLEU score points in the minimal and fluency tracks, respectively. The biggest performance increase was in Estonian — an 8.25 and 4.97 GLEU score increase for the minimal and fluency tracks, respectively. Notably, Estonian is not one of the pre-trained languages in Aya-Expanse. 

With the OmniGEC dataset, the model quality is more uniform: for AYA-Expanse, the lowest GLEU score improved by 8.26 points (minimal), but decreased by 3.05 GLEU points (fluency) on Icelandic track. Except for Icelandic, previously underperforming and unknown languages gained the most significant performance increase in both tracks. Gemma-3 scores improved by 4.99 (minimal) and 0.54 (fluency) GLEU scores. Both models outperformed the leading paragraph-based editing model in the MultiGEC competition (LLaMA-3-8B) when compared using the mean GLEU score.

Due to the cost and time considerations, Gemma-3 was trained only on one epoch with LoRA for all linear layers for both tracks. Gemma-3 took almost a day to complete a single epoch on a single A100 (40GB) GPU with packing and batching, whilst Aya-Expanse completed three training epochs within the same 24-hour window on the same GPU before hitting the plateau. Interestingly, Gemma-3 trained just for one epoch on OmniGEC and MultiGEC data outperformed Aya-Expanse in both tracks, although Aya-Expanse was more than 3 points ahead in the baseline performance for the minimal track. We hypothesize that such performance gain is due to Gemma-3 having more parameters and pre-trained language coverage, like for Latvian (GLEU increased by 26.14 points, compared to the baseline Gemma-3), Icelandic (up by 3.83 points), and Czech (up to 4.16 points). As we can observe, Gemma-3 benefits more than Aya-Expanse from extensive fine-tuning with a larger dataset, like OmniGEC.

For Icelandic, our results may not be directly comparable with those of MultiGEC participants, as we limited the number of generated tokens during inference to 1,600. This limitation did not impact any other languages; Icelandic test samples were longer than test samples in other languages, averaging at 1,000-3,000 tokens per essay. This hard cut might severely impact our performance in this language. Therefore, we leave further examination for future work.


 Refer to Table \ref{tab:config_comparison} (Appendix~\ref{sec:appendix-training-setup}) for the base hyperparameters used for Aya-Expanse and Gemma-3 models. For more details on the experiments, training, and model setup, refer to our GitHub \footnote{\url{https://github.com/r-kovalch/omnigec-models}}.


\subsection{Ablation Study}
\label{sec:ablation}

Although the same trend of uniform quality increase holds for both Aya-Expanse and Gemma-3, as we add more and more data, some individual languages oscillate in gained or lost performance, like Ukrainian fluency with the Aya-Expanse model, which bumped to 77.10 GLEU score (best score for paragraph-based edits) with MultiGEC+Wiki but lowered with the addition of the Reddit-MultiGEC dataset to 75.82 GLEU. This effect may be due to quality and structure variations of data per language in WikiEdits-MultiGEC and Reddit-MultiGEC. The same bump is present in Latvian for the Aya-Expanse model; however, Latvian gained more performance on Gemma-3, reaching 80.19 GLEU with even better results for MultiGEC+Wiki — 81.54 GLEU (best score for paragraph-based edits). On the other hand, for Estonian, the change is purely incremental for both models, with Gemma-3 achieving the state-of-the-art results using MultiGEC+Wiki+Reddit on Estonian minimal edits track. See Table \ref{tab:ft-ukr}.

Interestingly, for Gemma-3 the MultiGEC+Wiki track yields the best performance: 0.6 and 3.14 more GLEU points than MultiGEC+Wiki+Reddit for minimal and fluency tracks, respectively. Individual performance for some languages is also better with MultiGEC+Wiki than MultiGEC+Wiki+Reddit, e.g., Latvian increased by 1.35 GLEU points. We suppose that this performance increase is due to this track being trained for three more epochs as Wiki corpora is nearly 10 smaller than Reddit. That shows, that both models, although yielding good performance, are still undertrained — for both MultiGEC+Wiki and MultiGEC+Wiki+Reddit experiments with Gemma-3 we didn't reach the plateau. We leave further exploration to future work.

We suppose that differences like this are due to Ukrainian, a mid-resource language, being pre-trained on Aya-Expanse and potentially Gemma-3, in contrast to Estonian and Latvian, low-resource languages not supported by Aya-Expanse and with unknown support by Gemma-3. Estonian and Latvian benefited more from a large corpus of synthetic data than Ukrainian.

\section{Conclusions}

In this research, we presented the OmniGEC collection of multilingual silver-standard GEC corpora. We found that including more silver-grade training data improves accuracy in multilingual GEC. We demonstrated the performance increase by training Aya-Expanse (8B) and Gemma-3-12B-IT models on MultiGEC and OmniGEC datasets, which yielded the best results for paragraph-based editing models outperforming previous leaders trained solely on MultiGEC data. Aya-Expanse (8B), being a smaller model with fewer excess languages, adapted more easily to the multilingual GEC but has its limitations, like fewer relevant pre-training languages. These limitations can be addressed through fine-tuning on large-scale datasets in the target languages. Gemma-3-12B-IT, a larger model, despite having more parameters, yielded worse results than Aya-Expanse when trained solely on a small golden GEC dataset but after adding a large silver dataset for fine-tuning, outperformed Aya-Expanse and established a new paragraph-based editing SOTA score.

We publish OmniGEC and processing pipelines to open-source and expect OmniGEC to be continuously updated with new data, growing both in new samples and languages. The Reddit-MultiGEC and WikiEdits-MultiGEC subcorpora can be continuously updated with corrections. Together with our exploratory work, these resources aim to facilitate new developments in multilingual GEC with new models, approaches, and techniques.

In future work, we plan to further research multilingual GEC by assessing more models, sentence-based editing, which yielded better results in the MultiGEC-2025 shared task, and preference optimization methods, like DPO \cite{dpo}, made possible in this task with prepared human-in-the-loop scores in OmniGEC. On top of that, the ablation studies will be an important area for future research: (a) more thorough research on data quantity versus quality with UberText-GEC, which includes nearly 10 times more language data than Reddit-MultiGEC for the Ukrainian case study, and (b) per-language LoRA adapters to unveil the cross-language relationships, if any. Finally, we expect the UberText-GEC case study to trailblaze research toward the SOTA Ukrainian GEC model in both paragraph-based and sentence-based editing. We expect all these methods to easily adapt to other languages, improving multilingual GEC scores.

\section{Limitations}

We acknowledge the following limitations of our study:
\begin{itemize}
    \item OmniGEC covers only eleven languages, leaving aside the vast linguistic diversity.
    \item Human annotation feedback was collected only for the Ukrainian language, which makes it difficult to assess the quality of synthetically generated corrections for other languages and allows training a preference model only for Ukrainian.
    \item We used proprietary models for synthetic correction generation, which may impact the reproducibility of the approach.
    \item Due to time and cost restrictions, we trained Gemma-3-12B-IT only for one epoch and limited our research to two open-source multilingual LLMs.
\end{itemize}

\section{Ethical Considerations}

For Reddit-MultiGEC, we collected posts from publicly available subreddits and utilized the OpenAI content moderation API to filter out potentially harmful and offensive texts, as this data is later used for LLM fine-tuning and may impact model performance in unpredictable ways. Unfortunately, we do not have qualitative estimates on how well the moderation API works for the target eleven languages.

Additionally, we did not estimate the level of misinformation and biases in the multilingual Reddit posts.

\section{Acknowledgments}

We express our gratitude to the volunteers, students, and lecturers from the National Technical University "Kharkiv Polytechnic Institute" who joined and promoted our annotation project: Mariia Shvedova, Anna Pospekhova, Myron Prokopenko, Arsenii Lukashevskyi, Veronica Moroz, Olha Tochylina, Sofiia-Tereza Onysko, Ksennia Lyzhna, Tamila Krashtan, Nataliia Sheremett, Kateryna Astafyeva, Yurii Petrov, Maryna Vozikova, Andri Ruda, Anna Khuhaieva, and others. We also thank the Faculty of Applied Sciences of the Ukrainian Catholic University for providing the computing resources and OpenAI API access. We thank Oleksandr Skurzhanskyi, Applied Research Scientist at Grammarly, and the reviewers of this work for their thoughtful and thorough ideas, comments, and critique, which were imperative for conducting this study. 



\clearpage

\appendix
\twocolumn
\section{WikiEdits-MultiGEC}
\label{sec:appendix-wikiedits-multigec}
\subsection{Data Source Examples}
\begin{figure}[h]

\begin{subfigure}[b]{0.99\linewidth}
    \centering
    \includegraphics[width=\linewidth]{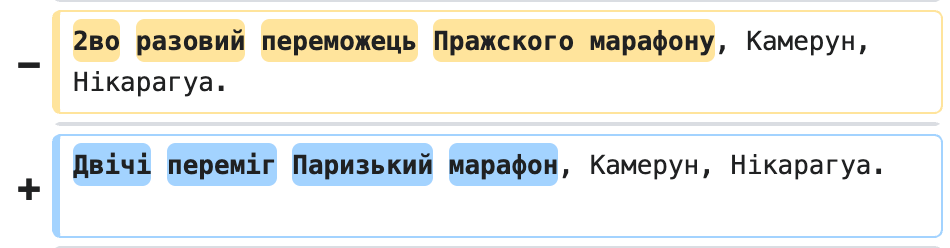}
    \caption{}
    \label{fig:correction-fig1}
\end{subfigure}
\hfill
\begin{subfigure}[b]{0.99\linewidth}
    \centering
    \includegraphics[width=\linewidth]{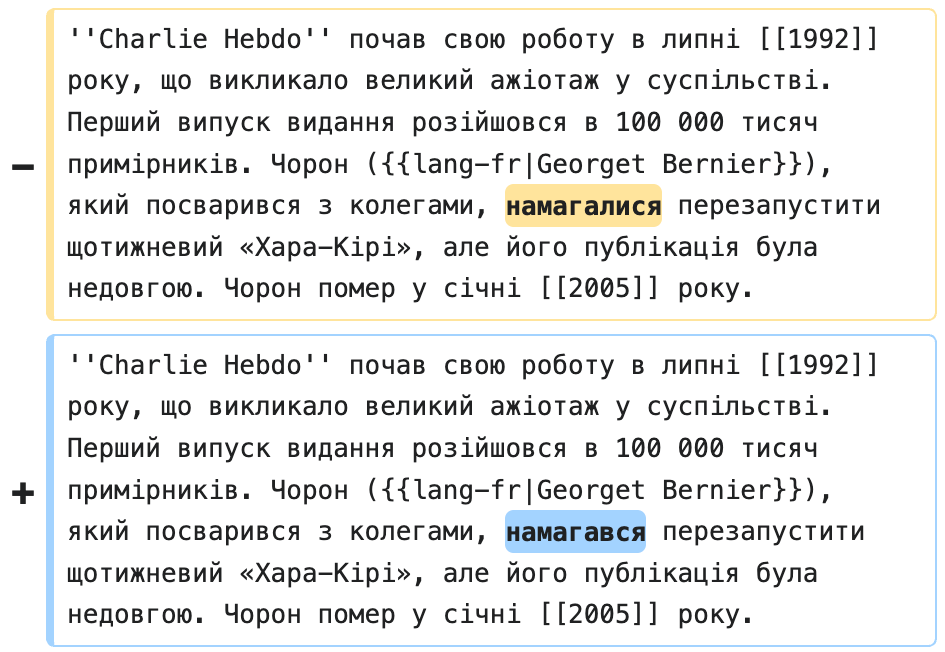}
    \caption{}
    \label{fig:correction-fig1}
\end{subfigure}
\caption{The examples of an edit diff from Wikipedia UI. The yellow(-) and blue(+) denote the removed and added text, respectively. (a) — example of an edit, (b) — example of a simple error correction.}
\label{fig:correction-fig}
\end{figure}

\subsection{Dataset Statistics}

\begin{table}[ht]
  \centering
  \footnotesize
  \begin{tabular}{lrrr}
    \hline
    \textbf{Language} & \textbf{\# pages} & \textbf{\# edits‑all} & \textbf{\# edits} \\
    \hline
    English          & 5,003 & 12,465 & 6,807 \\
    Italian          & 2,398 &  6,024 & 3,726 \\
    Ukrainian        & 1,409 &  5,126 & 3,092 \\
    German           & 1,706 &  4,672 & 2,380 \\
    Czech              &   447 &  1,114 &   698 \\
    Swedish            &   216 &    585 &   363 \\
    Greek              &   134 &    492 &   256 \\
    Estonian           &    39 &    126 &    79 \\
    Slovene            &    26 &    108 &    43 \\
    Latvian            &    20 &     75 &    33 \\
    Estonian &     0 &      0 &     0 \\
    \hline
  \end{tabular}
  \caption{Dataset creation steps: \# pages — pages with edits; \# edits-all — all edits from each page; \# edits — edits after filtering.}
  \label{tab:wiki-table}
\end{table}

\subsection{Data Filtering}
We applied the following filtering steps:

\begin{itemize}
\item We excluded samples shorter than 50 characters as they often represent unstructured or incomplete text fragments.
\item We excluded samples with more than 10 corrections as these generally signify extensive modification of the original text.
\item We excluded samples beginning with special characters (==, !, |, etc.,) as they usually denote Wikipedia-specific sections, tags, or formatting.
\item All samples were cleaned from custom Wikipedia formatting, such as referral links, citations, code tags, language-specific tags, etc.
\end{itemize}

\onecolumn
\section{Dataset Comparison}
\label{sec:appendix-datasets-comparison}
\begin{figure}[H]
  \includegraphics[width=\textwidth]{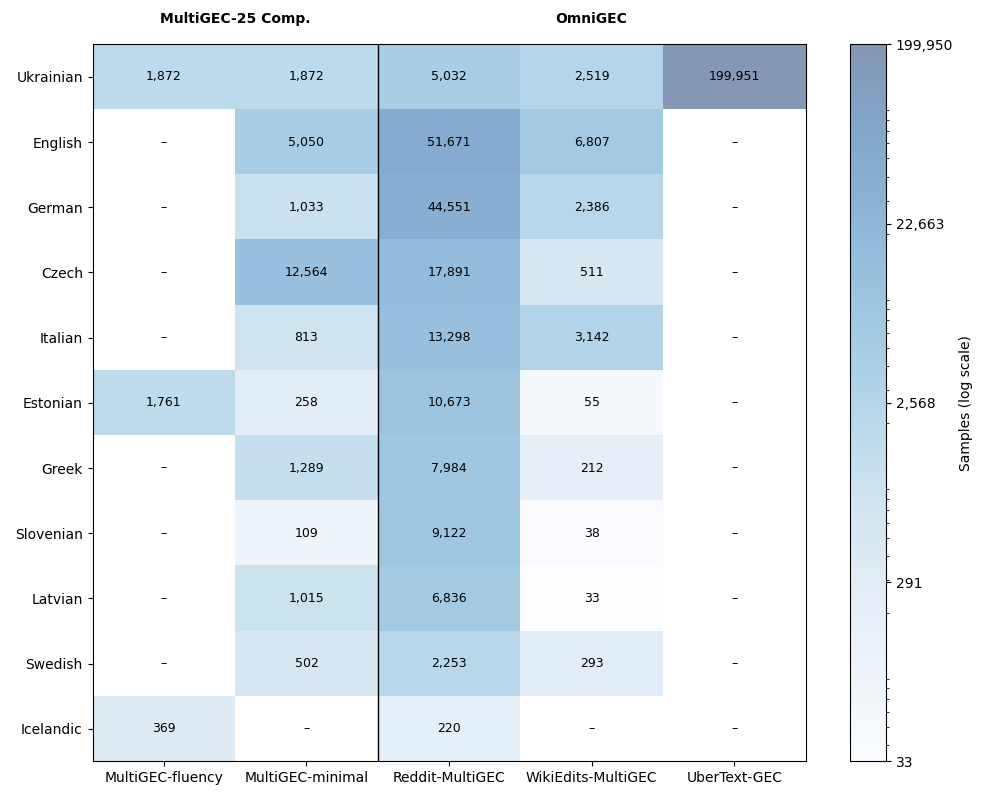}
  \caption[width=\textwidth]{Number of samples in the multilingual golden (MultiGEC‑25) and silver (OmniGEC) GEC datasets. Data was split 80\%/10\%/10\% into train/validation/test sets per language.}
  \label{fig:corpus_data}
\end{figure}
\clearpage
\begin{figure}[H]
  \includegraphics[width=\textwidth]{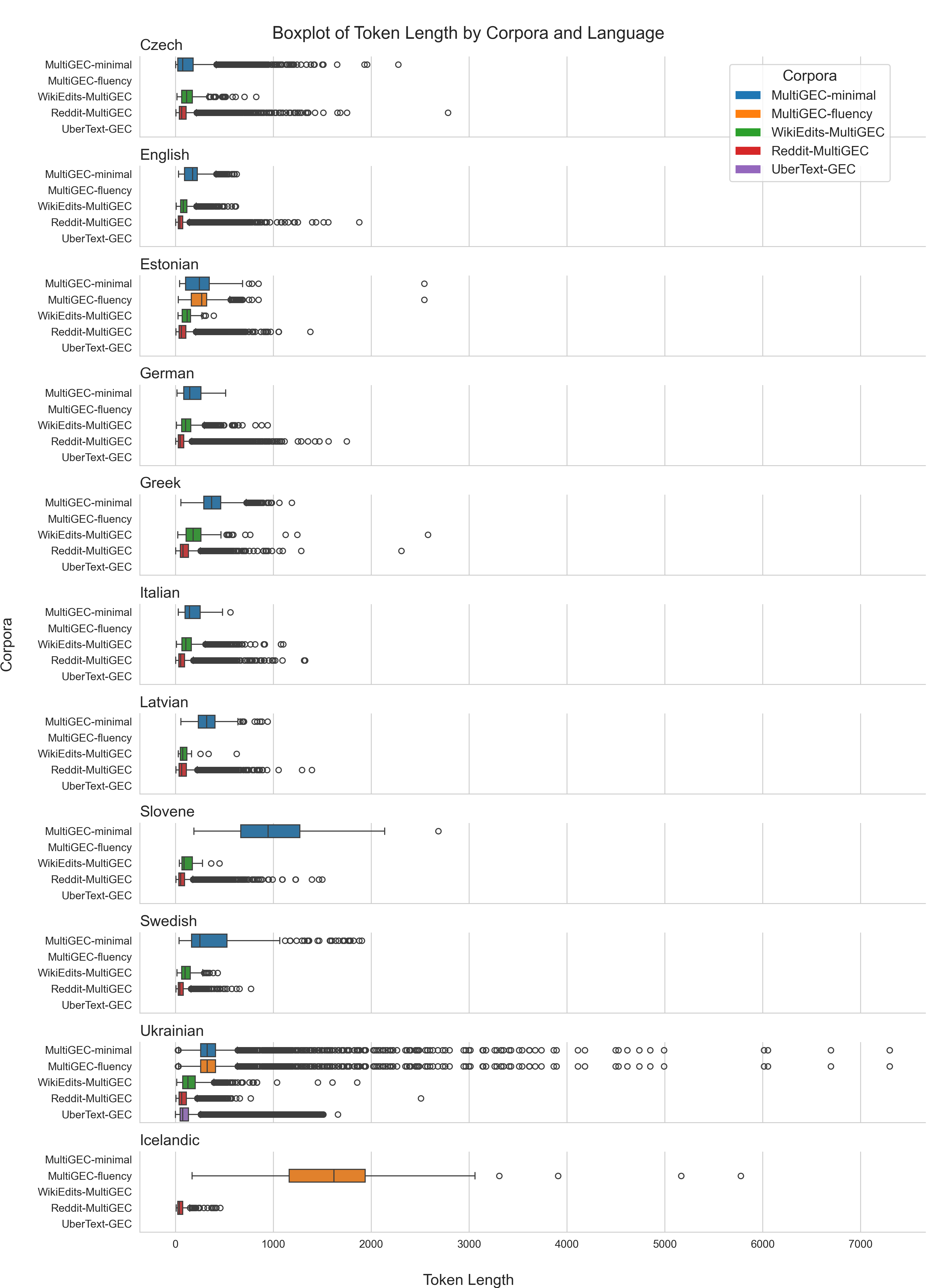}
  \caption[width=\textwidth]{Token‐length distributions by corpus and language for golden (MultiGEC-2025) and silver (OmniGEC) GEC datasets. We used the GPT-4o-mini tokenizer for assessing the length of the datasets.}
  \label{fig:token-len}
\end{figure}
\clearpage
\section{WikiEdits-MultiGEC Error Analysis}
\label{sec:appendix-error-analyis}

\begin{figure}[H]

\begin{subfigure}[b]{0.99\textwidth}
    \centering
    \begin{tcolorbox}[colback=white!95!gray,
                 colframe=black,
                 boxrule=0.5pt,
                 arc=4mm,  
                 width=\linewidth,
                 left=2mm, right=2mm, top=1mm, bottom=1mm]
    \selectlanguage{ukrainian}
\textbf{Text:} Норвегію на літніх Олімпійських іграх 2000 року, які проходили в Сіднеї, представляли {\color{red}93} спортсмени (44 чоловіків та 49 жінок) у {\color{red}15} видах спорту. Прапороносцем на церемонії відкриття Олімпійських ігор був бігун Вебйорн Родаль

\rule{\linewidth}{0.5pt}

\textbf{Correction}: Норвегію на літніх Олімпійських іграх 2000 року, які проходили в Сіднеї, представляли {\color{green}97} спортсмени (44 чоловіків та 49 жінок) у {\color{green}12} видах спорту. Прапороносцем на церемонії відкриття Олімпійських ігор був бігун Вебйорн Родаль

\rule{\linewidth}{0.5pt}

\textbf{Translation}: Norway was represented at the 2000 Summer Olympics in Sydney by 93 athletes (44 men and 49 women) in 15 sports. The flag bearer at the opening ceremony of the Olympic Games was runner Webjorn Rodal
    \end{tcolorbox}
    \caption{}
    \label{fig:error-analysis-a}
\end{subfigure}

\begin{subfigure}[b]{0.99\textwidth}
    \centering
    \begin{tcolorbox}[colback=white!95!gray,
                 colframe=black,
                 boxrule=0.5pt,
                 arc=4mm,  
                 width=\linewidth,
                 left=2mm, right=2mm, top=1mm, bottom=1mm]
    \selectlanguage{ukrainian}
\textbf{Text:} При взаємодії з гідроксиламіном утворює {\color{red}оксим}, який під дією оцтового ангідриду перетворюється на ацильований гідроксинітрил.

\rule{\linewidth}{0.5pt}

\textbf{Correction}: При взаємодії з гідроксиламіном утворює {\color{green}оксин}, який під дією оцтового ангідриду перетворюється на ацильований гідроксинітрил.

\rule{\linewidth}{0.5pt}

\textbf{Translation}: When it reacts with hydroxylamine, it forms oxime, which is converted to acylated hydroxynitrile under the action of acetic anhydride.
    \end{tcolorbox}
    \caption{}
    \label{fig:error-analysis-b}
\end{subfigure}

\begin{subfigure}[b]{0.99\textwidth}
    \centering
    \begin{tcolorbox}[colback=white!95!gray,
                 colframe=black,
                 boxrule=0.5pt,
                 arc=4mm,  
                 width=\linewidth,
                 left=2mm, right=2mm, top=1mm, bottom=1mm]
    \selectlanguage{ukrainian}
\textbf{Text:} Економічне благо — це товари й послуги, що є результатом доцільної діяльності людини.

\rule{\linewidth}{0.5pt}

\textbf{Correction}: Економічне благо — це товари й послуги, що є результатом доцільної діяльності людини. {\color{green}Вони створюються для задоволення людських потреб і вимагають витрат ресурсів, часу та зусиль.}

\rule{\linewidth}{0.5pt}

\textbf{Translation}: An economic good is goods and services that result from a person's reasonable activity.
    \end{tcolorbox}
    \caption{}
    \label{fig:error-analysis-c}
\end{subfigure}

\begin{subfigure}[b]{0.99\textwidth}
    \centering
    \begin{tcolorbox}[colback=white!95!gray,
                 colframe=black,
                 boxrule=0.5pt,
                 arc=4mm,  
                 width=\linewidth,
                 left=2mm, right=2mm, top=1mm, bottom=1mm]
    \selectlanguage{ukrainian}
\textbf{Text:} Із <math>{ a over b } = { c over d }</math> слідує (помножимо ліву і праву частину рівності на {\color{red}b)}:

\rule{\linewidth}{0.5pt}

\textbf{Correction}: Із <math>{ a over b } = { c over d }</math> слідує (помножимо ліву і праву частину рівності на {\color{green}"b")\:}

\rule{\linewidth}{0.5pt}

\textbf{Translation}: From <math>{ a over b } = { c over d }</math>, it follows (multiply the left and right sides of the equality by b:
    \end{tcolorbox}
    \caption{}
    \label{fig:error-analysis-d}
\end{subfigure}
\hfill

\caption{Error Analysis for the WikiEdits-MultiGEC dataset. Examples of errors: (a) Information updates; (b) Domain knowledge; (c) Distortion of context; (d) Data errors. All translations were performed using the DeepL service.}
\label{fig:error-analysis}
\end{figure}

\onecolumn
\section{Training Results}
\label{sec:appendix-training-results}
\begin{figure}[H]
  \includegraphics[width=\textwidth]{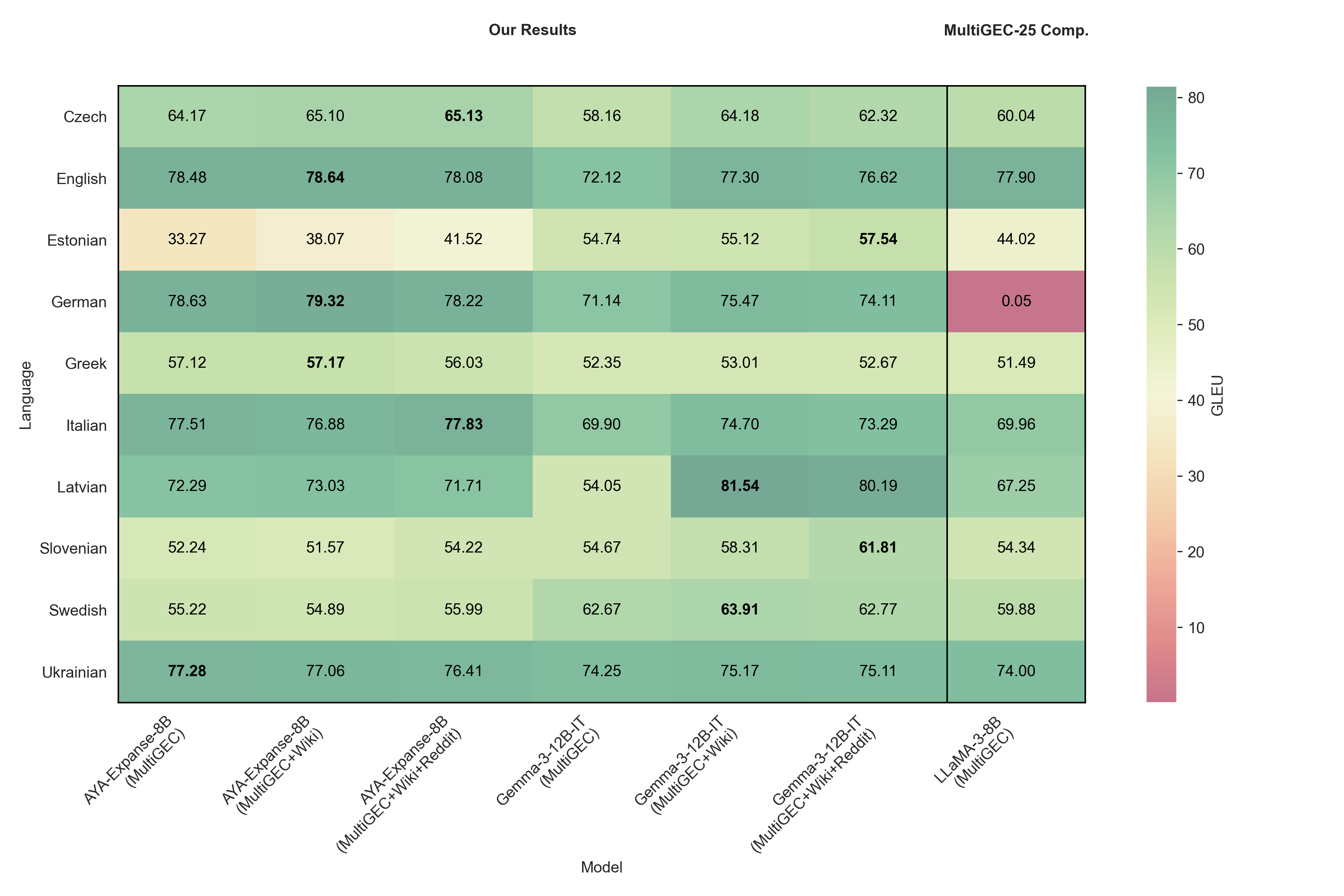}
  \caption{Face-to-face comparison of paragraph-based GEC models fine-tuned on the MultiGEC and OmniGEC datasets across all languages for the minimal track.}
  \label{fig:face2face-minimal}
\end{figure}
\begin{figure}[H]
  \includegraphics[width=\textwidth]{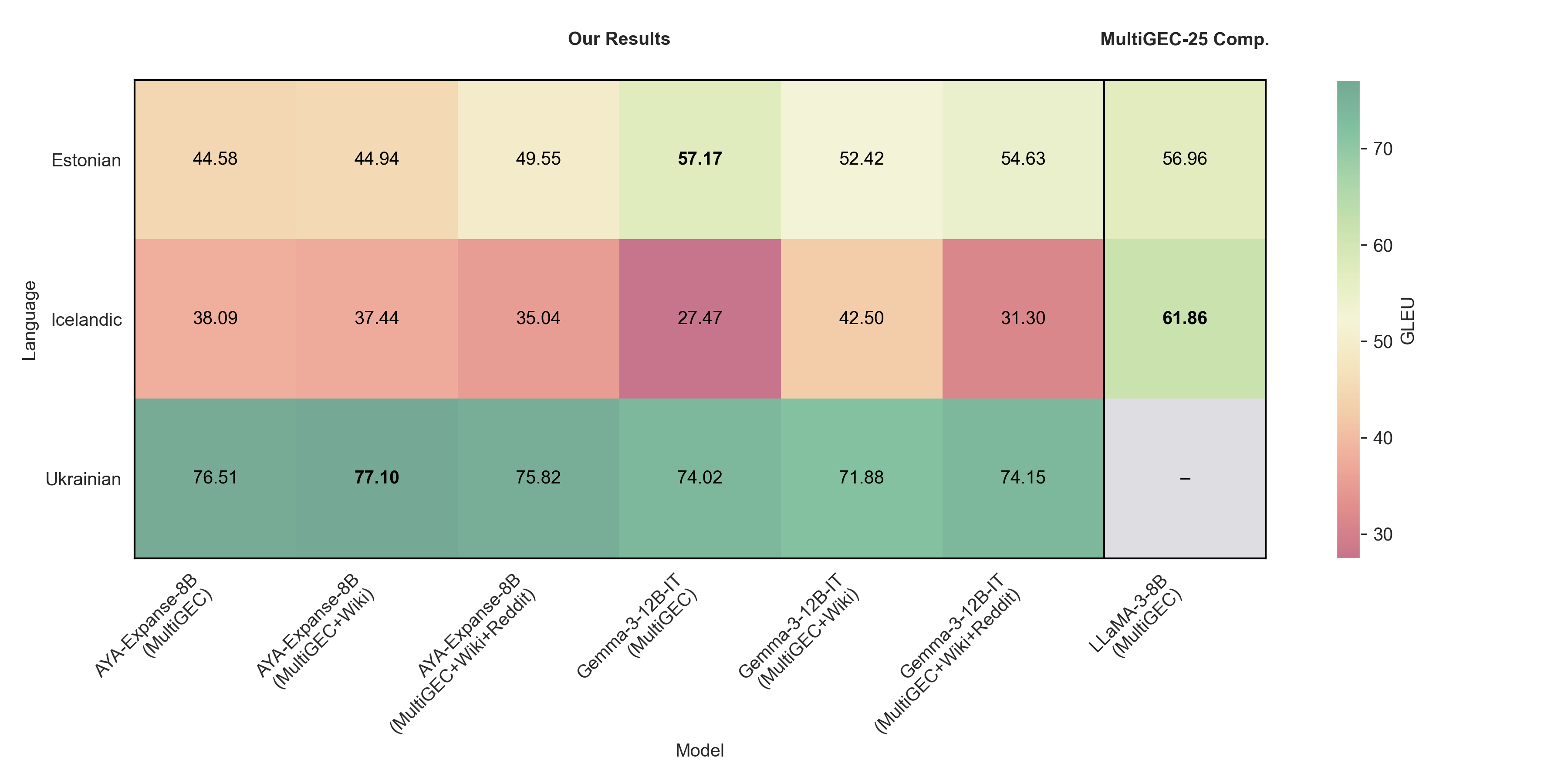}
  \caption{Face-to-face comparison of paragraph-based GEC models fine-tuned on the MultiGEC and OmniGEC datasets across all languages for the fluency track.}
  \label{fig:face2face-fluency}
\end{figure}

\onecolumn
\section{Training Setup}
\label{sec:appendix-training-setup}
\begin{table}[H]
\footnotesize
  \centering
  \begin{tabular}{lllcccccc}
    \toprule
    \textbf{Model} & \textbf{AYA-Expanse-8B} & \textbf{Gemma-3-12B-IT} \\
    \midrule
    \multicolumn{3}{l}{\textbf{Inference}} \\
    \midrule
    temperature          & 0.3  & 1.0  \\
    top\_p               & 0.75 & 0.95 \\
    top\_k               & 0    & 64   \\
    max\_new\_tokens     & 1600 & 1600  \\
    \midrule
    \multicolumn{3}{l}{\textbf{Training}} \\
    \midrule
    num\_train\_epochs              & 12 & 7  \\
    per\_device\_train\_batch\_size & 7  & 4  \\
    per\_device\_eval\_batch\_size  & 2  & 2  \\
    gradient\_accumulation\_steps   & 8  & 8  \\
    gradient\_checkpointing         & true  & true  \\
    optim                           & paged\_adamw\_32bit   & adamw\_torch\_fused  \\
    save\_steps                     & 100 & 100 \\
    logging\_steps                  & 10  & 10  \\
    learning\_rate                  & 3e-5 & 3e-5 \\
    weight\_decay                   & 0.0 & 0.0 \\
    max\_grad\_norm                 & 1.0 & 1.0 \\
    fp16                            & false & false \\
    bf16                            & true  & true  \\
    warmup\_steps                   & 50 & 70  \\
    group\_by\_length               & false & false \\
    lr\_scheduler\_type             & cosine & cosine \\
    report\_to                      & wandb & wandb \\
    eval\_strategy                  & steps & steps \\
    save\_strategy                  & steps & steps \\
    metric\_for\_best\_model        & eval\_loss & eval\_loss \\
    greater\_is\_better             & false & false \\
    save\_total\_limit              & 1 & 1 \\
    load\_best\_model\_at\_end      & true & true \\
    eval\_steps                     & 25 & 25 \\
    \midrule
    \multicolumn{3}{l}{\textbf{Early Stopping}} \\
    \midrule
    early\_stopping\_patience & 75  & 200 \\
    \midrule
    \multicolumn{3}{l}{\textbf{LoRA}} \\
    \midrule
    lora\_alpha     & 128 & 128 \\
    r               & 64  & 64  \\
    bias            & none & none \\
    task\_type      & CAUSAL\_LM & CAUSAL\_LM \\
    target\_modules & q\_proj,\,v\_proj,\,k\_proj,\,o\_proj,\,gate\_proj,\,up\_proj & all‑linear \\
    modules\_to\_save & default & lm\_head,\,embed\_tokens \\
    \bottomrule
  \end{tabular}
  \caption{Configuration of inference, training, early‐stopping, and LoRA base settings for AYA‑Expanse‑8B and \ Gemma‑3‑12B‑IT. For individual experiments, some parameters may differ. For details refer to our GitHub.
  }
  \label{tab:config_comparison}
\end{table}
\clearpage
\twocolumn

\end{document}